Original Article

# Intra-YOLO: A Small Object Detection Model for Caries and Molar-Incisor Hypomineralization in Intraoral Photography Based on Transfer Learning with Reinforcement Learning

Po-Lun Chwang[1,*], Po-Yu Chang[2,3,*], Wen-Liang Lin[1], Tung-Sheng Wu[1], Min-Ching Wang[2,3,+], Yun-Chien Cheng[1,+]

[1] Department of Mechanical Engineering, College of Engineering, National Yang Ming Chiao Tung University, Hsin-Chu, Taiwan

[2] Taipei Medical University Hospital, Taipei, Taiwan

[3] Wan Fang Hospital, Taipei Medical University, Taiwan

*: the authors contribute equally to this work

+: the authors contribute equally to this work

Correspondence author:

Name: Yun-Chien Cheng, E-mail: yccheng@nycu.edu.tw

Min-Ching Wang, E-mail: 110071@w.tmu.edu.tw

**Running Title:** Intra-YOLO for Caries and MIH Lesions in Intraoral Photography

## Abstract

**Objectives** This study developed a computer-aided diagnosis (CAD) system for detecting caries and molar-incisor hypomineralization (MIH) in intraoral photographs. These lesions share similar appearances, making clinical differentiation challenging, especially given their small size and variability in imaging conditions.

**Materials and Methods** We propose Intra-YOLO, a deep learning model optimized for small lesion detection. Based on YOLOX, it integrates an enhanced feature pyramid network, attention mechanisms, and sliced inference. Transfer learning with knowledge distillation from a cropped-image teacher model is combined with a reinforcement learning–based selection mechanism to prioritize informative regions.

**Results** On a clinical dataset, Intra-YOLO achieved mAP 23.1 and mAP50 47.8, outperforming other object detection models. The integration of attention, scale-aware distillation, and slicing significantly improved small lesion recognition.

**Conclusions** Intra-YOLO enables accurate, efficient detection of caries and MIH, potentially reducing diagnostic variability and supporting timely, evidence-based treatment planning in dentistry.

**Clinical Relevance** This study provides a computer-aided method to support clinicians in identifying dental caries and molar-incisor hypomineralization from intraoral images, with the potential to reduce diagnostic subjectivity, bias and facilitate earlier clinical decision-making.

**Keywords** Deep learning, Caries, Molar-Incisor Hypomineralization, Small object detection, Transfer learning, Sliced inference

## Introduction

Molar-incisor hypomineralization (MIH) is a qualitative developmental defect affecting enamel and dentin. This condition is common among children worldwide, impacting approximately 10 to 40 % of the global pediatric population[1-3]. (Prevalence and distribution of demarcated opacities and their sequelae in permanent 1st molars and incisors in 7 to 13-year-old Brazilian children, Prevalence of molar incisor hypomineralization (MIH) in Singaporean children) According to the standards provided by the European Academy of Pediatric Dentistry, nearly one-third of teeth affected by MIH can progress to more severe issues, such as post-eruptive breakdown and tooth sensitivity[4,5]. Additionally, teeth with MIH are significantly more susceptible to caries, making the prevention and treatment of these lesions particularly challenging.

Although MIH and caries share certain similarities in clinical presentation, their causes and treatment methods differ significantly. MIH is a congenital or developmental defect, whereas caries results from demineralization caused by organic acids produced by bacteria. Accurate diagnosis of developmental defects is essential for effective clinical differentiation. Misdiagnosis can lead to ineffective or inappropriate treatment[6]. Due to their highly similar clinical presentations, distinguishing between these lesions poses a significant challenge for general dentists, often resulting in diagnostic inaccuracies[7]. Even when observing multiple lesions over the entire oral cavity, visual examinations among different dentists can vary significantly, relying heavily on individual experience[8]. This reliance makes accurate, objective diagnosis difficult, complicating the selection of treatment methods.

To address these challenges, this study designed and developed a computer-aided diagnosis (CAD) aimed at providing more precise and consistent diagnostic results. The system uses intraoral photography to capture full-mouth images during clinical examinations. These images are then processed through the automated diagnostic system, which extracts lesion features of caries and MIH. Based on these features, the system accurately classifies and localizes the lesions, generating predictive bounding boxes. This approach may help dentists handle visually similar lesions with greater precision and consistency, serving as an objective diagnostic aid.

Previous studies on molar-incisor hypomineralization (MIH) and caries detection in intraoral photography remain limited, with most research focusing on radiographic imaging [9–13]. Existing methods often use single-tooth images or lack comparative analysis with caries, limiting clinical applicability and increasing misdiagnosis risk due to similar lesion appearances. For instance, Jule et al. [14] developed a ResNeXt-101–based classification model using single-tooth images to categorize MIH lesions of varying severity and normal teeth. Intraoral photography poses additional challenges, including saliva contamination, lighting variability, and the extremely small size of lesions—over 78% of annotated lesions occupy less than 0.58% of the image area—making them prone to omission during feature extraction (Fig. 1) [15]. To address these gaps, we propose a deep learning–based model optimized for small lesion detection in intraoral images, aiming to improve diagnostic accuracy and support clinical decision-making. In conclusion, we present Intra-YOLO, a small object detection framework and computer-aided diagnostic system for identifying caries and molar-incisor hypomineralization (MIH) in intraoral photography. By integrating attention mechanisms, an enhanced feature pyramid network, transfer learning with knowledge distillation, reinforcement learning, and sliced inference, the framework addresses challenges related to lesion size, image variability, and noise. This system analyzes routine intraoral photographs and highlights suspected caries and MIH lesions, providing clinicians with an objective reference to support diagnosis and facilitate earlier treatment planning.

# Methods

## Participants

This study collected images from patients undergoing routine dental treatments at Taipei Municipal Wanfang Hospital between 2022 and 2025. Intraoral photographs were captured primarily using a Canon EOS R camera with an EF 100 mm lens, from various angles, and included patients with healthy teeth, caries, or developmental defects. A total of 822 images were collected, with annotations for 3,583 caries and 4,443 MIH lesions. The dataset was randomly split into training and validation sets in an 8:2 ratio. The training set included 657 images with 2,874 caries annotations and 3,669 MIH annotations, while the validation set contained 165 images with 709 caries annotations and 774 MIH annotations. The study was approved by the Taipei Medical University Hospital Institutional Review Board (TMU-JIRB #N202212082).

## Data Preprocessing

To enhance the detection model's ability to extract features from small lesions, the training set was divided into patches of approximately 640×640 pixels. Images without annotated bounding boxes were removed to avoid background imbalance. A total of 6,417 cropped images were retained, containing 5,119 caries and 7,122 MIH bounding boxes, effectively increasing the number of valid training samples.

## Model Overview

The proposed Intra-YOLO framework combines transfer learning and reinforcement learning to enhance small lesion detection in intraoral photography (Fig. 2). It adopts a teacher–student paradigm in which both teacher and student models are based on an improved YOLOX architecture, referred to as S-YOLO. The teacher S-YOLO is trained on cropped high-resolution image patches and incorporates attention mechanisms (SSM)[16], an enhanced feature pyramid network (S-PAFPN), and sliced inference[17] to maximize small-lesion feature extraction. The S-PAFPN module enhances multi-scale feature fusion by combining high-resolution details from the largest feature map with semantic information from early backbone layers, preserving fine-grained lesion features. An integrated attention mechanism further suppresses noise and background interference, improving small lesion detection accuracy.

After training, the teacher's predictions are used as pseudo-labels to guide the student S-YOLO, which is trained on full-size images to integrate localized and global features. An adaptive thresholding strategy, combined with reinforcement learning, is employed to decide whether a predicted bounding box should be distilled. The reinforcement learning agent makes this decision based on lesion size, the teacher model's confidence score, and the student model's prediction uncertainty measured by entropy loss. This targeted, scale-aware distillation enables the student to focus on high-value learning cases, improving detection accuracy for small and variable-size lesions while maintaining computational efficiency.

# Results

## Evaluation Metrics

Model performance was primarily evaluated using mean Average Precision (mAP) across both lesion classes. Additional metrics included mAP50 at a fixed IoU of 0.5, and APs, $AP_m$, and $AP_L$ to assess detection performance for small, medium, and large lesions. As no public intraoral photography dataset is available, all evaluations were conducted on the clinical dataset collected from Taipei Municipal Wanfang Hospital.

## Detection and Visualize Results

On the clinical intraoral photography dataset, S-YOLO achieved 22.0% mAP and 44.2% mAP50, Intra-YOLO improved performance to 23.1% mAP and 47.8% mAP50 (Table 1). Figs. 3-5 present detection outputs of Intra-YOLO on intraoral images from multiple views, including frontal anterior, occlusal, and lateral perspectives. Caries are indicated by blue bounding boxes, and MIH lesions by yellow bounding boxes. The model accurately identified lesions across various tooth regions, including challenging areas such as occlusal junctions and posterior teeth, demonstrating robust performance in multi-view intraoral imagery.

## Discussion

The high diagnostic accuracy of Intra-YOLO has the potential to support clinical dentists in distinguishing molar–incisor hypomineralization (MIH) from dental caries, which may reduce misdiagnosis and facilitate more appropriate treatment planning. The superior performance of Intra-YOLO in detecting small dental lesions can be attributed to the integration of several key strategies for small object detection. Existing approaches—such as scale adjustment via Feature Pyramid Networks (FPN) and Spatial Pyramid Pooling (SPP)[18-20], anchor-based sampling[21-23], feature imitation[24-26], and post-processing with sliced inference—have each demonstrated benefits in specific contexts but also have inherent limitations when applied to intraoral imagery. In our framework, we enhanced multi-scale feature fusion through an improved pyramid network (S-PAFPN) and attention mechanisms inspired by state-space modeling, enabling better localization of fine-grained lesion features. Transfer learning with knowledge distillation allowed effective transfer of small-lesion features from a patch-trained teacher model to a full-image student model, while reinforcement learning dynamically selected high-value predictions for distillation based on reward function . Together, these components contributed to the observed gains in mAP and mAP50, particularly for small and scale-variant lesions.

### Impact of Reinforcement Learning on Model Performance

Reinforcement learning (RL) played a key role in further improving the detection performance of Intra-YOLO. RL is well-suited for small object detection tasks that require discrete decision-making[27-29], such as adaptive region selection and scale adjustment, where static strategies often fail to capture sparse, localized features. Building on prior work demonstrating the stability and effectiveness of Proximal Policy Optimization (PPO) for multi-objective optimization, we incorporated PPO into the knowledge distillation process. In our design, the RL agent dynamically determines whether a bounding box should be distilled based on lesion size, the teacher model's confidence score, and the student model's prediction uncertainty.

Ablation experiments on dynamic IoU thresholds (φ) confirmed the value of this adaptive approach (Table 2). While looser static thresholds improved match rates, they risked introducing noisy supervision; conversely, stricter thresholds suppressed effective distillation for small objects. Our PPO-based learnable strategy achieved the highest overall performance, with mAP reaching 23.1% and mAP50 reaching 47.8%. Small object detection (mAPs) improved to 18.3%, with concurrent gains in $AP_m$ and $AP_L$, indicating consistent benefits across scales. These results highlight that scale-aware, confidence-guided distillation via RL enables more targeted and effective learning, particularly for small and scale-variant lesions.

**Comparison of End-to-End Models**

When compared with a range of end-to-end object detection models in Table 3, Intra-YOLO achieved the highest performance on intraoral photography. Two-stage models such as Faster R-CNN[30] and Transformer-based approaches (DETR[31], DAB-DETR[32]) showed lower mAP values (12.4%–16.7%), likely reflecting the limited dataset size and the challenges of small lesion detection in intraoral images. Single-stage detectors such as YOLOv5 and YOLOX performed better, with YOLOX combined with sliced inference reaching 19.7% mAP.

Our improved S-YOLO, which incorporates attention mechanisms and an enhanced feature pyramid network, achieved 22.0% mAP and 44.2% mAP50. The full Intra-YOLO framework, integrating S-PAFPN, SSM, and PPO-based scale-aware distillation, further improved performance to 23.1% mAP and 47.8% mAP50. These results indicate that the proposed design effectively addresses the detection of small and scale-variant lesions, outperforming both two-stage and single-stage baselines and demonstrating its clinical potential for accurate and efficient intraoral diagnostics.

## Conclusions

From a clinical perspective, this approach may assist dentists in more reliably identifying caries and molar-incisor hypomineralization, potentially reducing diagnostic variability, supporting earlier intervention, and improving patient outcomes. While further validation on larger and more diverse datasets is warranted, the results indicate that Intra-YOLO could serve as a valuable decision-support tool in dental practice.

In this study, the proposed S-YOLO model, integrating SSM, S-PAFPN, and sliced inference, achieved 22.0% mAP and 44.2% mAP50. Incorporating PPO-based, scale-aware knowledge distillation into the Intra-YOLO framework further improved performance to 23.1% mAP and 47.8% mAP50, representing gains of 4.7% and 8.5% over the baseline YOLOX. These results suggest that combining enhanced multi-scale feature fusion, attention mechanisms, and reinforcement learning–guided distillation can substantially improve the detection of small and scale-variant lesions in intraoral photography.

## Conflict of Interest

The authors declare that they have no conflict of interest.

## Ethics Approval

This study was approved by the Institutional Review Board of Taipei Medical University Hospital (TMU-JIRB #N202212082).

## Consent to Participate

Informed consent was obtained from all participants included in the study.

## Data Availability

The datasets generated and analyzed during the current study are not publicly available due to patient privacy but are available from the corresponding author on reasonable request.

**Table 1** Detection Results of Caries and MIH

| Model | mAP | mAP50 | mAP75 |
|---|---|---|---|
| **S-YOLO** | 22.0 | 44.2 | 18.5 |
| **Intra-YOLO** | 23.1 | 47.8 | 19.6 |

**Table 2** Performance comparison of static IoU thresholds and PPO-based adaptive distillation

| $\phi$ | ***mAP*** | ***mAP50*** | ***mAP75*** | ***mAPs*** | ***APm*** | ***mAPL*** |
|---|---|---|---|---|---|---|
| **X (IoU=0.5)** | 19.9 | 42.5 | 15.6 | 14.3 | 13.4 | 21.9 |
| **0.1** | 22.2 | 44.5 | 18.7 | 16.6 | 14.4 | 22.9 |
| **0.15** | 21.9 | 43.8 | 17.2 | 17.0 | 13.3 | 22.4 |
| **0.2** | 19.8 | 42.7 | 15.9 | 13.8 | 12.3 | 21.9 |
| **PPO** | **23.1** | **47.8** | **19.6** | **18.3** | **14.5** | **25.6** |

**Table 3** Comparison of End-to-End Models

| End-to-end model | | | |
|---|---|---|---|
| **Model** | **mAP** | **mAP50** | **mAP5** |
| **FRCNN** | - | 23.4 | - |
| **DETR** | 12.4 | 32.5 | 7.5 |
| **DAB-DATR** | 13.8 | 35.3 | 9.4 |
| **DAB-Deformable-DETR** | 16.7 | 38.3 | 12.1 |
| **YOLOv5** | 18.8 | 38.8 | 14.2 |
| **YOLOX** | 18.4 | 39.3 | 14.5 |
| **YOLOX+slicing** | 19.7 | 42.5 | 15.5 |
| **S-YOLO** | 22.0 | 44.2 | 18.5 |
| **Intra-YOLO** | **23.1** | **47.8** | **19.6** |

## Figure legends

**Fig 1** Area ratio of lesion in intaoral photography

**Fig 2** Overview of Intra-YOLO structure

**Fig 3** Lesion detection results under frontal view

**Fig 4** Lesion detection results under occlusal view of upper and lower jaws

**Fig 5** Lesion detection results under lateral view

**Fig. 1**

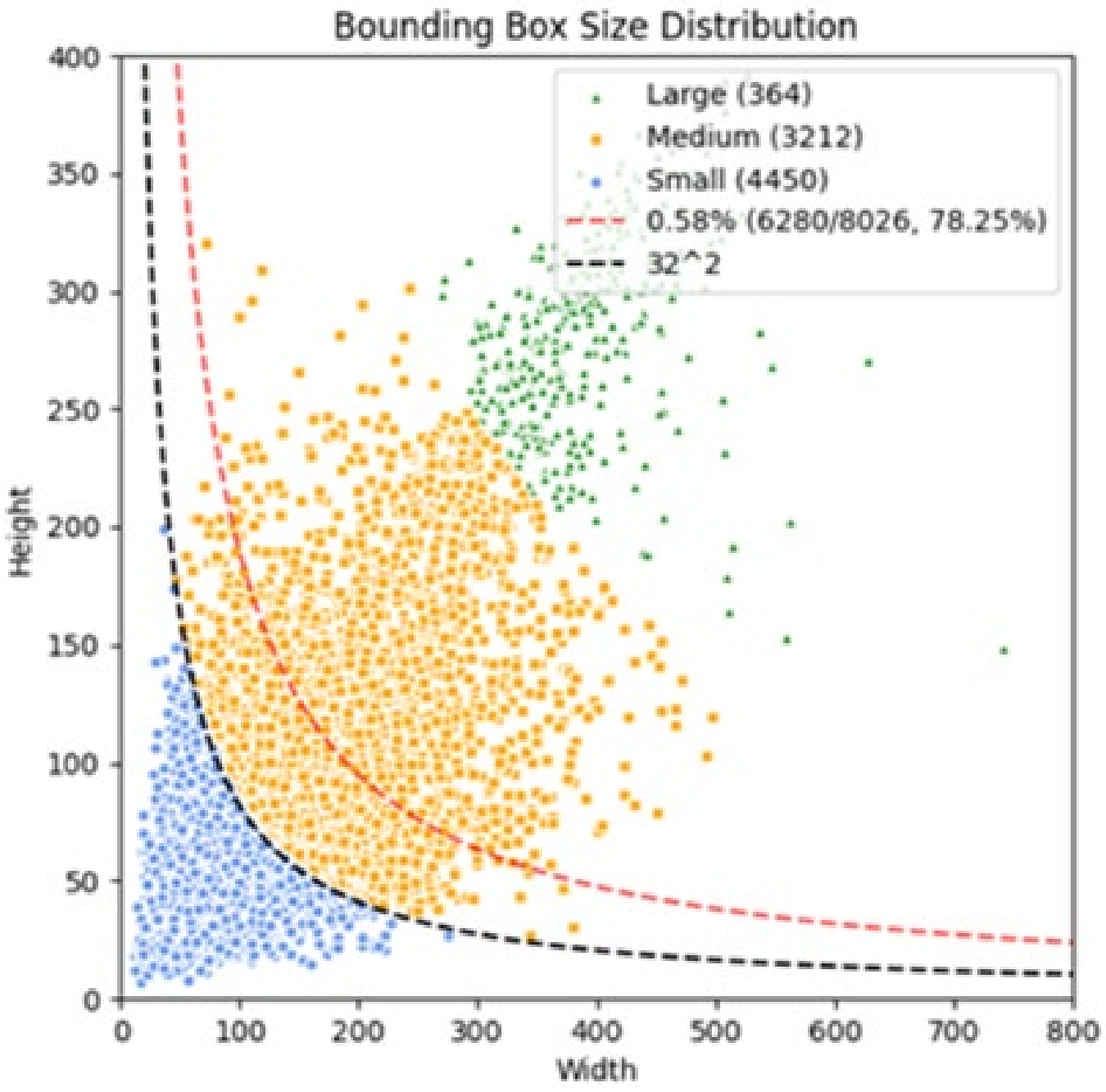
Bounding Box Size Distribution
Large (364)
Medium (3212)
Small (4450)
0.58% (6280/8026, 78.25%)
32^2
Height
Width
0
50
100
150
200
250
300
350
400
0
100
200
300
400
500
600
700
800

**Fig. 2**

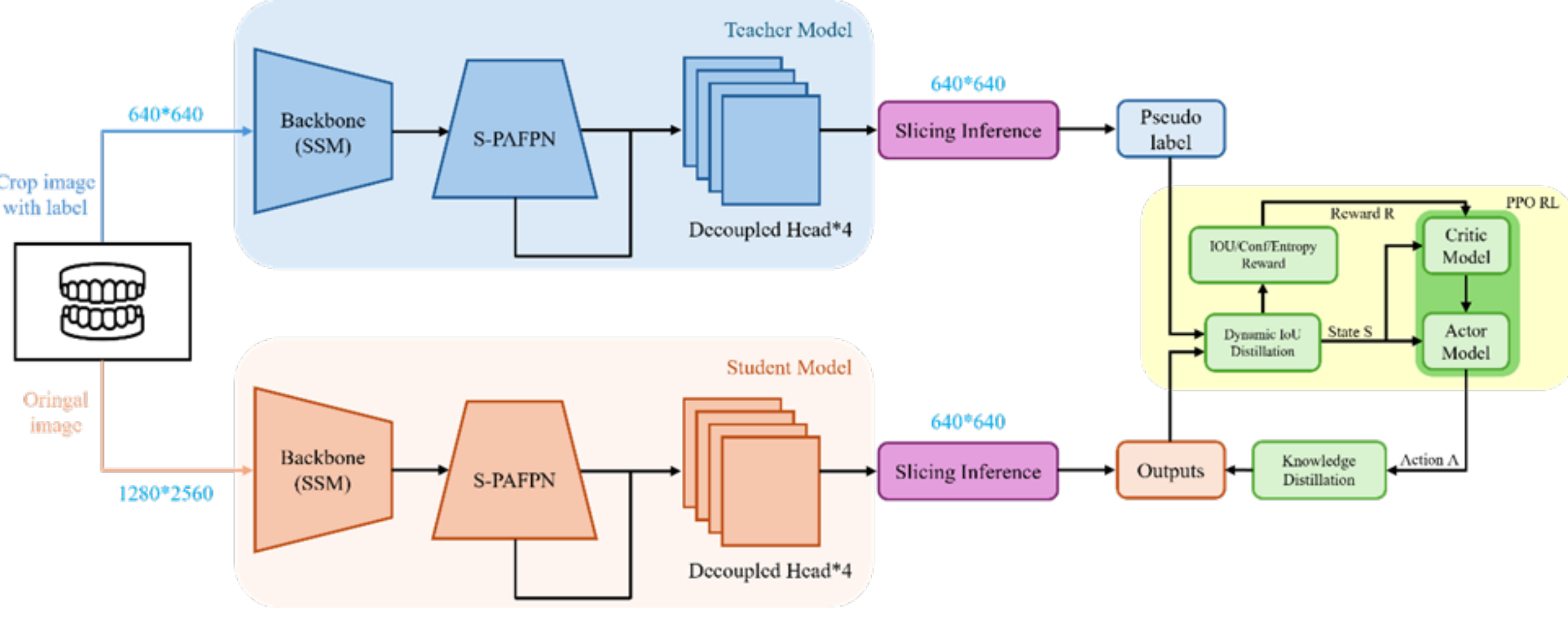
Teacher Model
640*640
Backbone (SSM)
S-PAFPN
Decoupled Head*4
640*640
Slicing Inference
Pseudo label
Crop image with label
Oringal image
Student Model
1280*2560
Backbone (SSM)
S-PAFPN
Decoupled Head*4
640*640
Slicing Inference
Outputs
PPO RL
Reward R
IOU/Conf/Entropy Reward
Critic Model
Dynamic IoU Distillation
State S
Actor Model
Knowledge Distillation
Action A

## Fig. 3

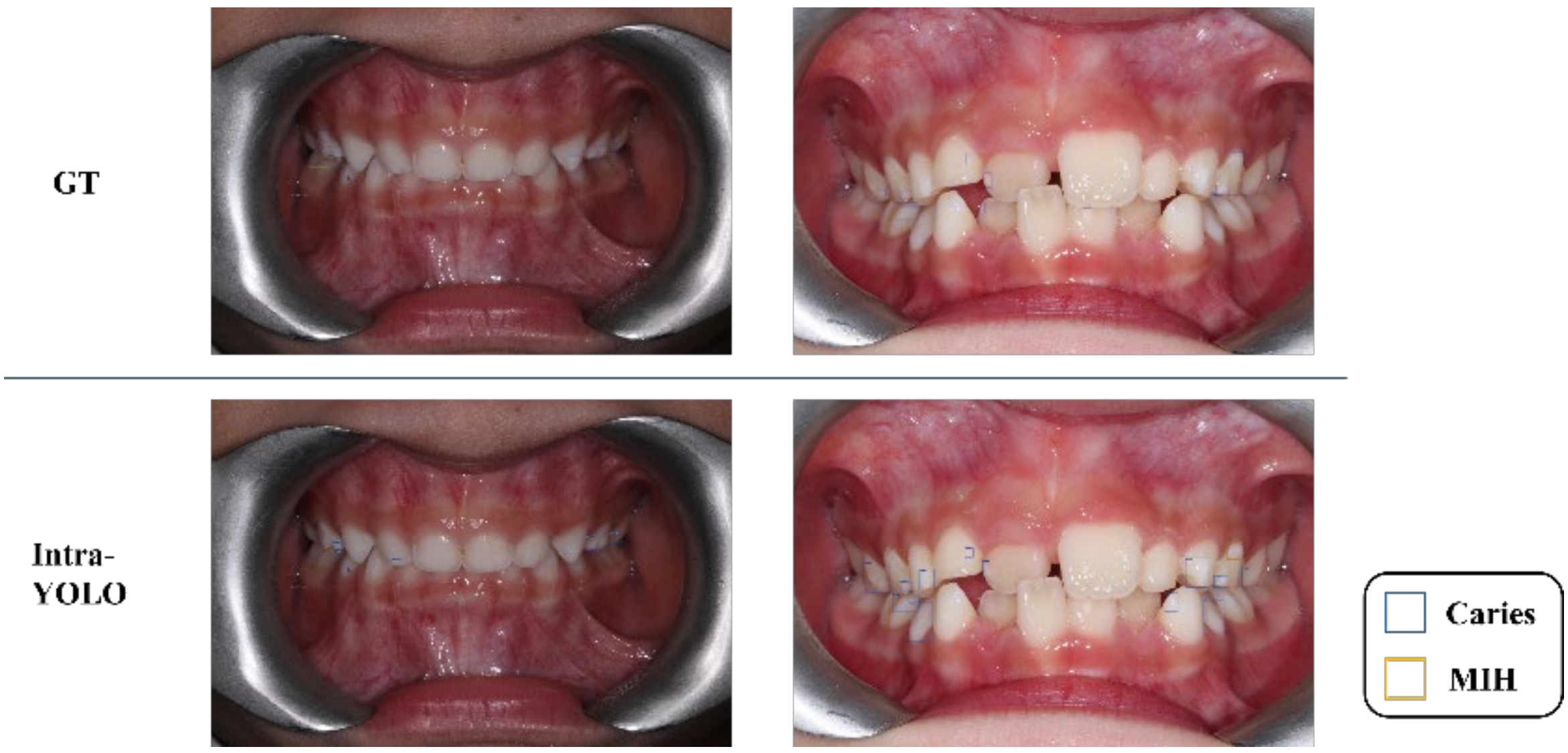

## Fig. 4

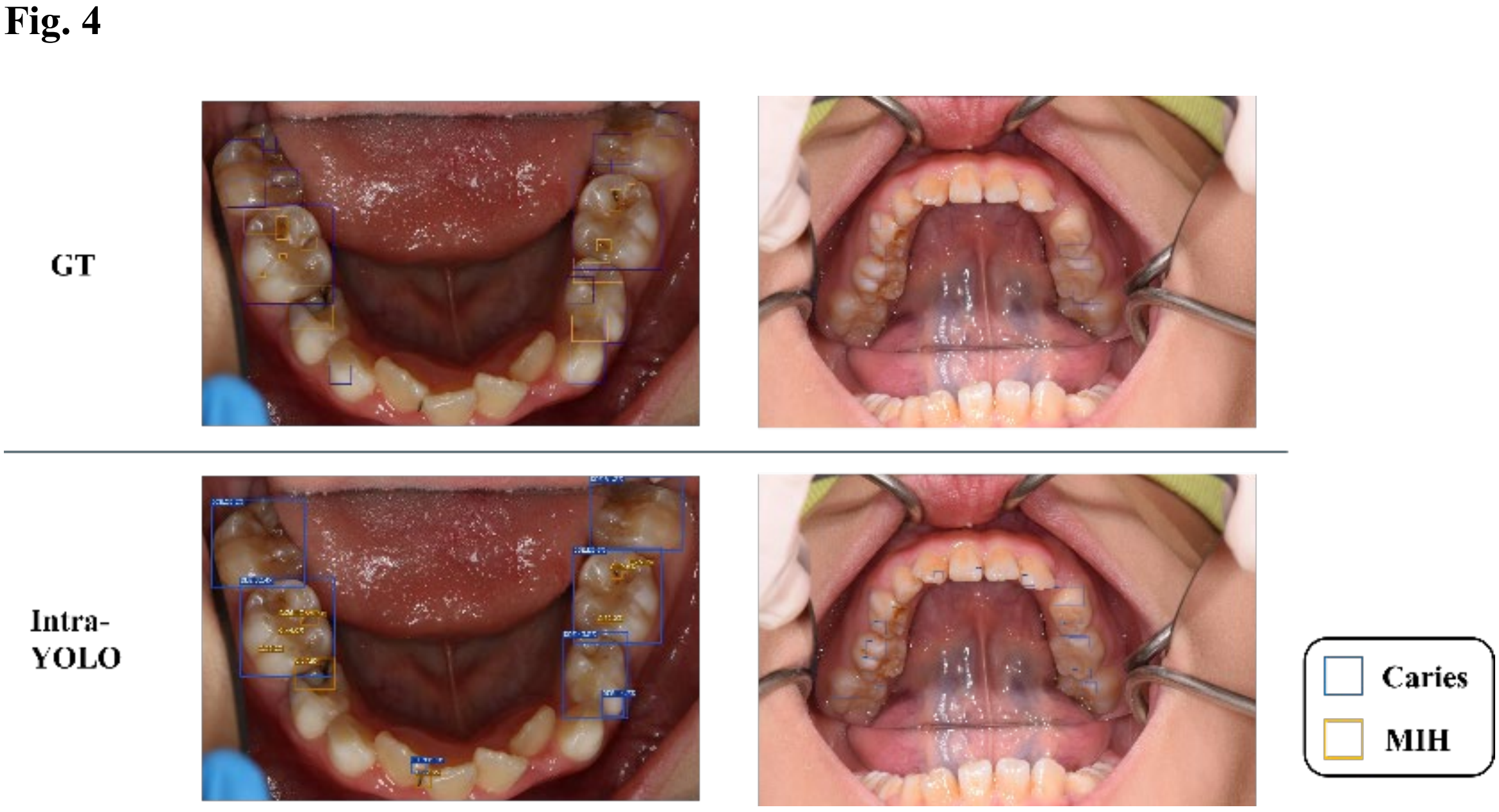

**Fig. 5**

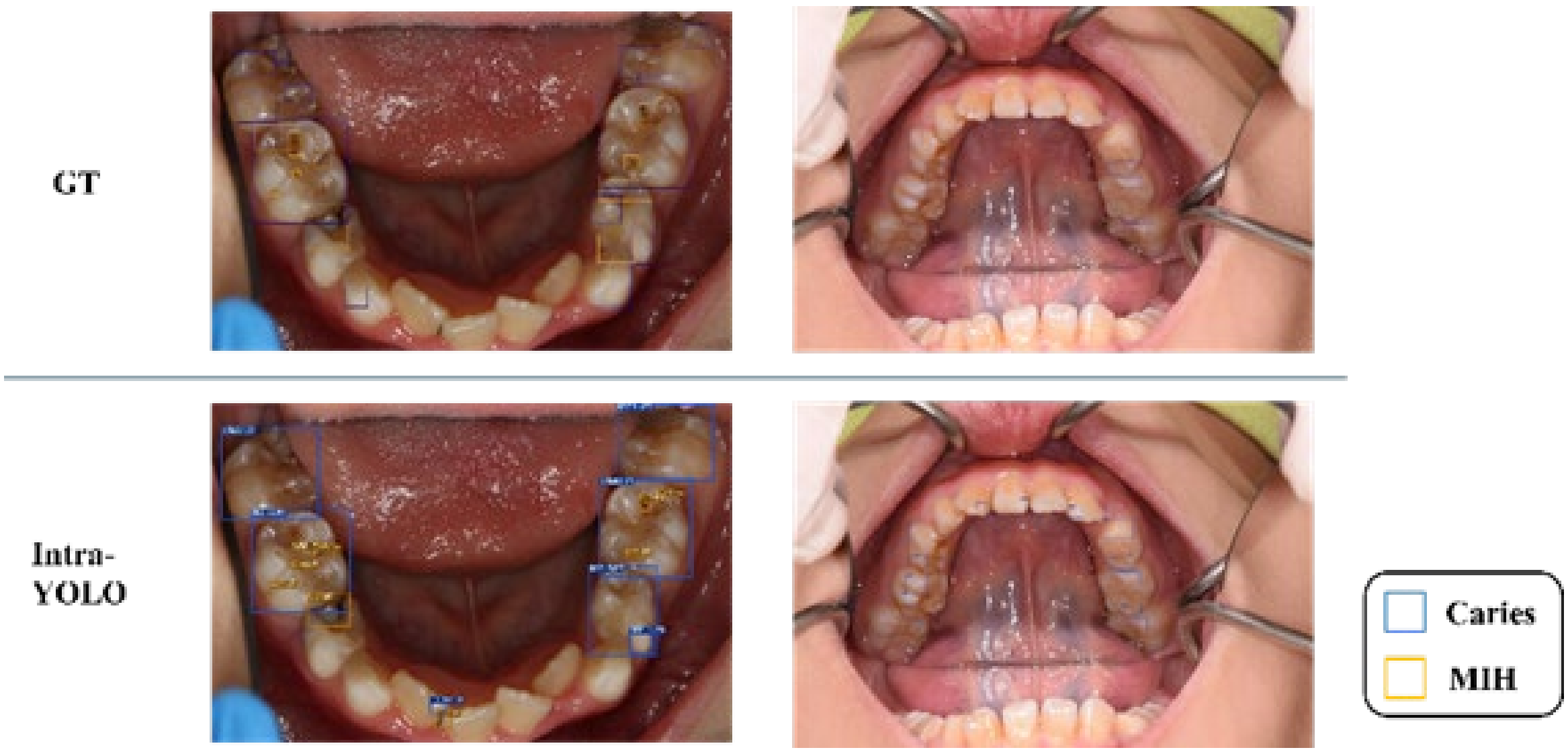